\documentclass{article}
\usepackage{bm}
\usepackage{spconf,amsmath,graphicx} \usepackage{booktabs}
\usepackage{color}

\usepackage{amsfonts}
\def\x{{\mathbf x}}

\title{Identification of Deep Breath While Moving Forward based on Multiple Body Regions and Graph Signal Analysis}
\name{Yunlu Wang$^1$, Cheng Yang$^2$, Menghan Hu$^{1,2,*}$, Jian Zhang$^1$, Qingli Li$^1$, Guangtao Zhai$^2$, Xiao-Ping Zhang$^3$}
\address{$^1$ Shanghai Key Laboratory of Multidimensional Information Processing, East China Normal University\\
$^2$ Key Laboratory of Artificial Intelligence, Ministry of Education\\
$^3$ Department of Electrical, Computer and Biomedical Engineering, Ryerson University\\
\thanks{This work is sponsored by the National Natural Science Foundation of China (No. 61901172), the China Postdoctoral Science Foundation funded project (No. 2020TQ0194), and the Science and Technology Commission of Shanghai Municipality (No. 19511120100).}
\thanks{$^*$Corresponding author: Menghan Hu (mhhu@ce.ecnu.edu.cn)}}
\begin{document}
%
\maketitle
\begin{abstract}
This paper presents an unobtrusive solution that can automatically identify deep breath when a person is walking past the global depth camera. Existing non-contact breath assessments achieve satisfactory results under restricted conditions when human body stays relatively still. 
When someone moves forward, the breath signals detected by depth camera are hidden within signals of trunk displacement and deformation, and the signal length is short due to the short stay time, posing great challenges for us to establish models. To overcome these challenges, multiple region of interests (ROIs) based signal extraction and selection method is proposed to automatically obtain the signal informative to breath from depth video. Subsequently, graph signal analysis (GSA) is adopted as a spatial-temporal filter to wipe the components unrelated to breath. Finally, a classifier for identifying deep breath is established based on the selected breath-informative signal. In validation experiments, the proposed approach outperforms the comparative methods with the accuracy, precision, recall and F1 of 75.5\%, 76.2\%, 75.0\% and 75.2\%, respectively. This system can be extended to public places to provide timely and ubiquitous help for those who may have or are going through physical or mental trouble.
\end{abstract}
\section{Introduction}
\label{sec:intro}
\begin{figure}[t]
\centering
\includegraphics[height=3.83cm,width=8.5cm]{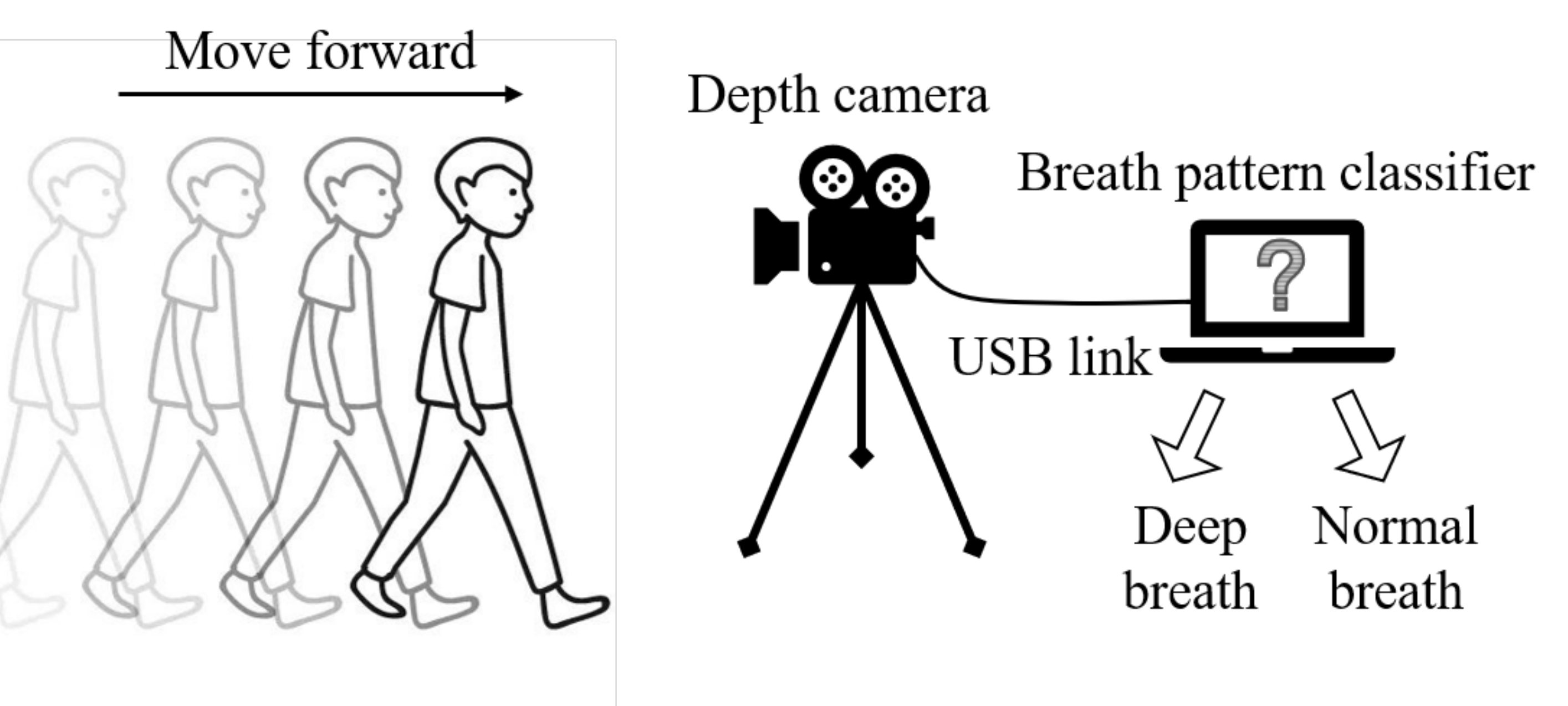}
\caption{Illustration of application scenario of our system. A person moves past the global depth camera, and the breath pattern classifier can infer whether the person has taken deep breaths without a stop at the camera.}
\label{experiments}
\end{figure}

In some scenarios, it would be significant to be able to measure breath as people move around. For example, in a shopping mall, if it is possible to detect the respiratory conditions of customers as they stroll through the mall, and the relevant departments will be able to provide immediate assistance for people who are going through physical or mental trouble. When human body stays relatively still, the non-contact breathing assessment has a satisfactory performance \cite{chen2018deepphys,cho2017deepbreath,wang2020unobtrusive,zhao2018noncontact,zhu2019vision,al2017real,janssen2015video}. These methods tend to fail when people move forward. Hence, we aim to find an unobtrusive solution to reduce the relative motionless restriction, and finally identify deep breath when a person moves forward. The application scenario of this solution is illustrated in \textbf{Fig. \ref{experiments}}.

To the best of our knowledge, there are no researches to assess respiratory state using a global camera when a person moves forward. In terms of in-head camera, researchers put a portable thermal imaging camera on the glass \cite{ruminski2016analysis} or fix it on the hat \cite{cho2017robust}, so they can measure breath during walking. However, the in-head camera may block the view, and affect user's respiratory state because he or she is already aware of being examined  \cite{lanfranchi1999prognostic}. 
Besides, these in-head or in-hand imaging architectures are suitable for personal usage, and they are difficult to be extended to the large-scale usage. 
Inspired by our previous research \cite{yang2017estimating}, we attempt to develop an algorithm to excavate the pure signals related to breath based on depth video when people walk past the global depth camera. 

Deep breath i.e., hyperpnea is characterized by increased depth of ventilation, and it is associated with stress, emotion and pulmonary infections \cite{RICARD201453}. We envision a non-contact approach that can detect deep breath on a large scale in public places, as long as someone walks past the global camera. Thus, timely and ubiquitous help can be provided to these people who may have emotional or physical problems. This unobtrusive assessment will not bring an additional burden to people, and it also can increase the level of service and safety in public places.

Two challenges inevitably exist in grabbing the deep breathing signal when a person walks past the global depth camera: 1) the signals associated with deep breath are hidden within signals of trunk displacement and deformation, and 2) since the human body's presence in front of the depth camera is brief, the effective recording time is very short, typically less than 10 seconds. Therefore, the contributions of this paper are as follows:

$\bullet$ We design an algorithm to generate the signals that account for the largest proportion of respiratory signals from multiple body regions, and purify them to obtain the signals informative to breath.

$\bullet$ We develop a graph signal analysis algorithm specific to the signals derived from depth video for filtering out components unrelated to breath.

$\bullet$ We establish a breath druing walking dataset which contains normal and deep breathing patterns to validate the proposed algorithm.

\section{Method}
\label{sec:method}
\begin{figure}[htb]
\centering
\includegraphics[height=9.39cm,width=8.6cm]{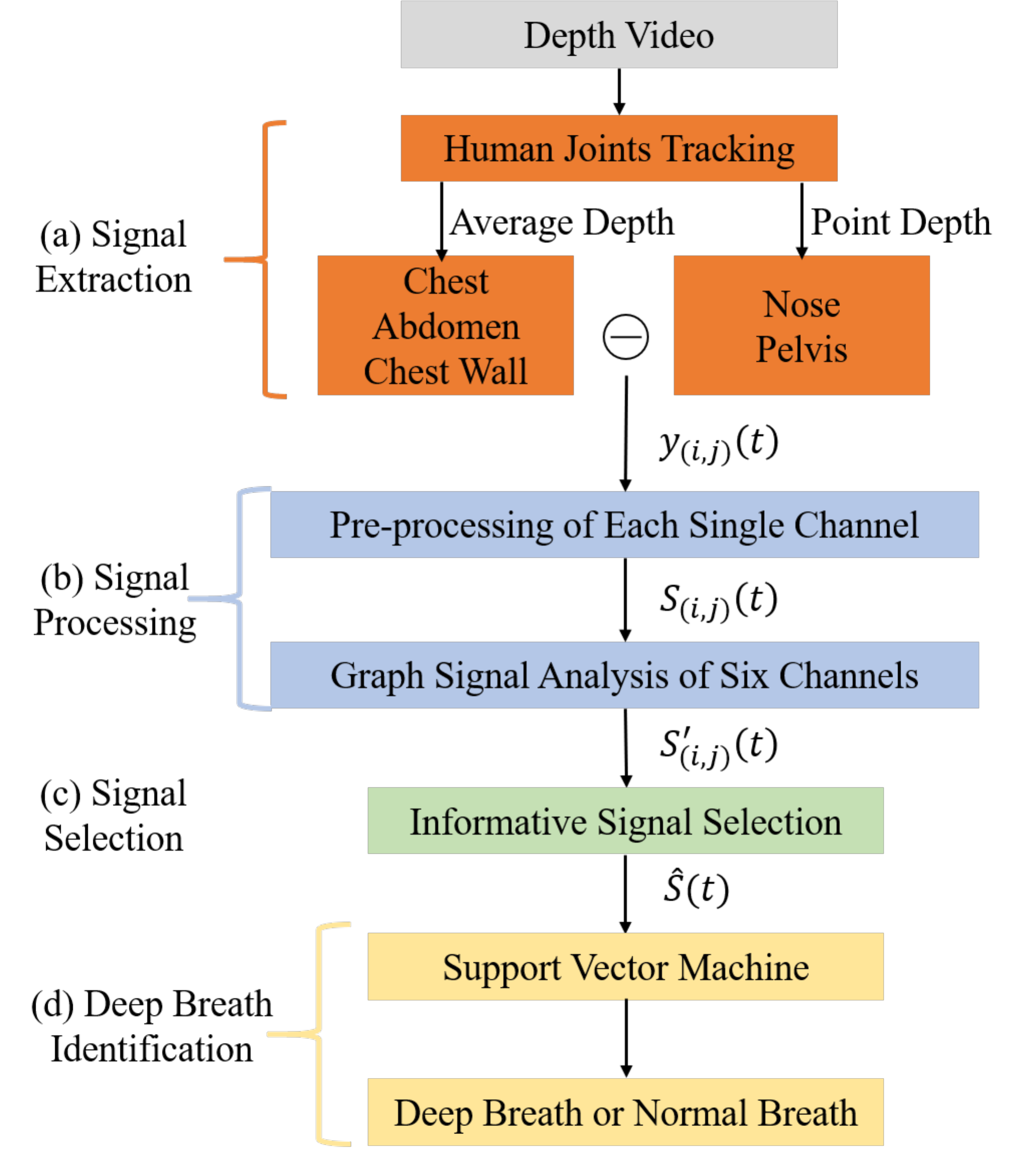}
\caption{Pipeline of the proposed non-contact deep breath identification from depth video. (a) Extract the relatively pure signals related to breath by subtracting the depth of nose or pelvis from the average depth of specific areas. (b) Pre-process each single channel using outlier removal, trend removal and band pass filter, then denoise all channels using GSA. (c) Select the most periodic signal for identification. (d) Establish deep breath identification model based on SVM.}
\label{pipeline}
\end{figure}

The pipeline of our method is shown in \textbf{Fig. \ref{pipeline}}. Firstly, we obtain the raw signal of six channels by multiple ROIs based signal extraction method: we automatically divide the chest and abdomen area into three parts viz. chest, abdomen and chest wall via human joints tracking, then we subtract the depth of nose or pelvis from the average depth of these three areas to produce one-dimensional (1D) temporal signal. Secondly, we pre-process each channel of the raw signal, then we adopt graph signal analysis (GSA) for spatial-temporal filtering. Thirdly, we choose the most suitable signal from six processed signals by informative signal selection method. Finally, a deep breath identification model is established using support vector machine (SVM).

\begin{figure}[htb]
\centering
\includegraphics[height=8.05cm,width=8.6cm]{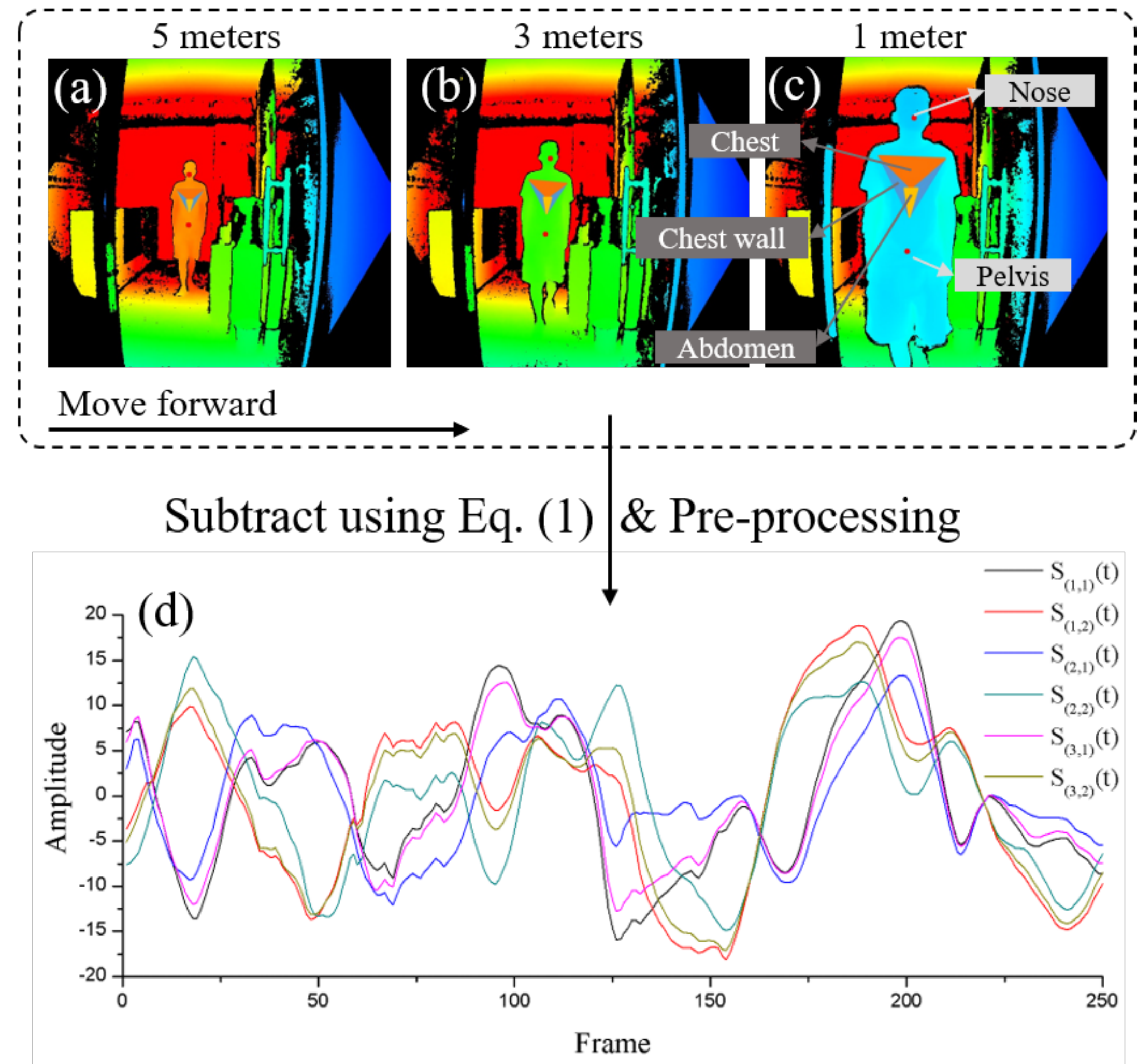}
\caption{One visualized example of the multiple ROIs based breath signal extraction when a person moves forward. 
Depth images during testing when the subject is 5 meters (a), 3 meters (b), and 1 meter (c) away from the depth camera. ROIs and stable points are labeled in (c). 
(d) We subtract the depth of one stable point from the average depth of one ROI using Eq. (1) to get the raw signal $y_{(i,j)}(t)$, then we pre-process each channel of the raw signal to get $S_{(i,j)}(t)$.}
\label{rois}
\end{figure}

\subsection{Multiple ROIs based respiratory signal extraction}
The inhalation and exhalation during breath cause the chest and abdomen to rise and fall \cite{kaneko2012breathing}, thus the breath signal can be extracted from depth value of those areas. Previous researches \cite{wang2020unobtrusive,prochazka2016microsoft} generally take the absolute depth as the raw signal.  

As people may have different breathing habits when they walk, we propose the multiple ROIs based respiratory signal extraction method. It includes three body regions related to breath and two stable points. As shown in \textbf{Fig. \ref{rois}}, based on human joints tracking, we take chest, abdomen and chest wall as the breath-related areas, and take pelvis and nose as the stable points. Then, we subtract the depth of one stable point from the average depth of one specific area to get one channel of the raw signal:
\begin{equation}
y_{(i,j)}(t) = \left[\frac{1}{n_{i}}\sum_{x,y\in ROI_{i}}D_{(x,y)}(t)\right] - D_{(x_{j},y_{j})}(t)
\end{equation}
where $D_{(x,y)}(t)$ represents the depth value at pixel $(x,y)$ in frame $t$; $n_{i}$ represents the number of pixels in the $i$th ROI; and $(x_{j},y_{j})$ represents the pixel of the $j$th stable point. In this research, $i=1,2,3$ and $j=1,2$.
Therefore, a total of six channels can be extracted from depth video.

\subsection{Graph signal analysis for spatial-temporal filtering}
Before GSA, we pre-process \cite{yang2017estimating, balakrishnan2013detecting} each channel of the raw signal $y_{(i,j)}(t)$. We replace outliers by linear interpolation, remove trend by least square method, and filter each channel to a passband of $ [0.167, 0.667]$ Hz by Butterworth filter.
We denote the pre-processed signal as $S_{(i,j)}(t)$.

Subsequently, the GSA technique \cite{shuman13,ortega18ieee} is adopted for spatial-temporal filtering of the signal extracted from multiple ROIs. 
GSA has been widely used in image processing field, such as denoising \cite{dinesh20193d} and contrast enhancement \cite{cheung18}.
In this work, we aim to obtain the denoised signal $\mathbf{x}=\{S_{(i,j)}'(t)\}$ given the six 1D temporal signal $\mathbf{y}=S_{(i,j)}(t)$.
Mathematically, we concatenate the six 1D temporal signal $S_{(i,j)}(t)$ into a vector $\mathbf{y}$.
We define an adjacency matrix $\mathbf{A}$ with following entries:
\begin{align}
w_{i,j}=
\left\{\begin{array}{l}
\exp\left\{-(\mathbf{f}_i-\mathbf{f}_j)^\top\mathbf{M}(\mathbf{f}_i-\mathbf{f}_j)\right\}, \;\;\mbox{if}\; i\neq j\\
0,\;\; \mbox{o.w.} 
\end{array}\right.
\end{align}
where $\mathbf{M}\in\mathbb{R}^{3\times3}$ is a metric matrix that measures the similarity among features of two samples, $\mathbf{f}_i\in\mathbb{R}^3$ denotes the sample feature, \textit{i.e.}, the horizontal and vertical pixel locations and pixel intensity of an observation.
We then define a degree matrix $\mathbf{D}=\sum_{i} w_{i,j}$.   
Therefore, a combinatorial graph Laplacian can be expressed as: $\mathbf{L}=\mathbf{D}-\mathbf{A}$.    

Now, we formulate a following \textit{Maximum a Posteriori} (MAP) problem:
\begin{equation}
\label{eq:gsp_obj}
\min_{\mathbf{x}} ||\mathbf{y}-\mathbf{x}||_2^2+\mu\x^\top\mathbf{L}\x    
\end{equation}
Eq. \eqref{eq:gsp_obj} has a closed-form solution $\mathbf{x}=(\mathbf{I}+\mu\mathbf{L})^{-1}\mathbf{y}$.
Since $\mathbf{L}\succeq0$, $\mathbf{I}+\mu\mathbf{L}\succ0$, and thus $\mathbf{I}+\mu\mathbf{L}$ is invertible and the solution of Eq. \eqref{eq:gsp_obj} is stable. We update the metric $\mathbf{M}$ that defines the feature distances in $\mathbf{L}$ via iterative gradient descent and positive definite (PD)-cone projection.  
Specifically,
we define $\mathbf{x}^\top\mathbf{L}(\mathbf{M})\mathbf{x}=Q(\mathbf{M})$, and thus the above iterative procedure is given by:    
\begin{equation}
\label{eq:proj}
\mathbf{M}_{t}=\mbox{Proj}(\mathbf{M}_{t-1}-\alpha_{t-1}\nabla Q(\mathbf{M}_{t-1}))
\end{equation}
where $\alpha$ is a step size that is initialised heuristically, increased by a small amount if the gradient descent yields a smaller objective and decreased by half otherwise, the gradient w.r.t. $M_{m,n}$ is given by:
\begin{equation}
\frac{\partial Q(\mathbf{M})}{\partial M_{m,n}}=-\sum_{i,j}
\Delta(\mathbf{f}_{ij})_m
\Delta(\mathbf{f}_{ij})_n
w_{i,j}
(x_i-x_j)^2
\end{equation}
where $\Delta \mathbf{f}_{ij}=\mathbf{f}_i-\mathbf{f}_j$.
To perform PD-cone projection, we first find the eigen-decomposition of
\begin{equation}
\mathbf{M}_{t-1}-\alpha_{t-1}\nabla Q(\mathbf{M}_{t-1})=\sum_i\lambda_i\mathbf{v}_i\mathbf{v}_i^\top
\end{equation}
We then define the projection:
\begin{equation}
\mathbf{M}_t=\sum_i\max\{\lambda_i,0\}\mathbf{v}_i\mathbf{v}_i^\top   
\end{equation}
We iteratively solve Eq. \eqref{eq:gsp_obj} and Eq. \eqref{eq:proj} until convergence. 

Based on the above analysis, we can filter out some components unrelated to breath, and finally get six denoised signals viz. $\mathbf{x}^*=\{S^{'}_{(i,j)}(t)\}$ in Eq. \eqref{eq:gsp_obj}.

\subsection{Informative signal selection}
After GSA, there still exist six signals extracted from different ROIs. As only one temporal signal is required for deep breath identification, we select the most periodic signal as the signal informative to breath. We adopt the quantification of periodicity in \cite{balakrishnan2013detecting}, which is originally used to select the most suitable eigenvector after Principal Component Analysis (PCA). This quantification specific to our research can be expressed as the following equation:
\begin{equation}
index = \frac{P_{f_{m}}+P_{2f_{m}}}{\sum_{i=0.167}^{0.667}{P_{i}}}
\end{equation}
where $f_{m}$ represents the frequency with maximal power in the passband of $[0.167, 0.667]$ Hz, and $P_{i}$ represents the power of frequency $i$. We then take the signal with largest $index$ as the signal informative to breath, and denote it as $\hat{S}(t)$.

\begin{table*}[ht]
\centering
\caption{Performance comparison of deep breath identification model on breath during walking dataset specific to different signal extraction and processing methods.}
\begin{tabular}{@{}cccccc@{}}
\toprule
Signal Extraction &
Signal Processing &
Accuracy(\%) &
Precision(\%) &
Recall(\%) &
F1(\%) \\ \midrule
\textbf{Mutiple ROIs \& Selection} &
\textbf{Bandpass \& GSA} &
  \textbf{75.5 $\pm$ 7.2} &
  \begin{tabular}[c]{@{}c@{}}\textbf{76.2 $\pm$ 8.0} \end{tabular} &
  \begin{tabular}[c]{@{}c@{}}\textbf{75.0 $\pm$ 10.2}\end{tabular} &
  \begin{tabular}[c]{@{}c@{}}\textbf{75.2 $\pm$ 7.6}\end{tabular} \\
Mutiple ROIs \& Selection &
  Bandpass &
  \begin{tabular}[c]{@{}c@{}}$66.7\pm8.7$\end{tabular} &
  \begin{tabular}[c]{@{}c@{}}$67.3\pm9.4$\end{tabular} &
  \begin{tabular}[c]{@{}c@{}}$66.4\pm10.3$\end{tabular} &
  \begin{tabular}[c]{@{}c@{}}$66.5\pm8.9$\end{tabular} \\
Single ROI &
  Bandpass \& GSA &
  \begin{tabular}[c]{@{}c@{}}$59.8\pm7.8$\end{tabular} &
  \begin{tabular}[c]{@{}c@{}}$58.5\pm6.9$\end{tabular} &
  \begin{tabular}[c]{@{}c@{}}$69.1\pm11.2$\end{tabular} &
  \begin{tabular}[c]{@{}c@{}}$63.0\pm7.7$\end{tabular} \\
Single ROI &
  Bandpass &
  \begin{tabular}[c]{@{}c@{}}$57.3\pm8.1$\end{tabular} &
  \begin{tabular}[c]{@{}c@{}}$56.8\pm7.7$\end{tabular} &
  \begin{tabular}[c]{@{}c@{}}$61.6\pm11.5$\end{tabular} &
  \begin{tabular}[c]{@{}c@{}}$58.8\pm8.6$\end{tabular} \\
PCA &
  Bandpass \& GSA &
  \begin{tabular}[c]{@{}c@{}}$44.1\pm7.1$\end{tabular} &
  \begin{tabular}[c]{@{}c@{}}$43.9\pm7.5$\end{tabular} &
  \begin{tabular}[c]{@{}c@{}}$43.0\pm12.1$\end{tabular} &
  \begin{tabular}[c]{@{}c@{}}$42.9\pm8.7$\end{tabular} \\
PCA &
  Bandpass &
  \begin{tabular}[c]{@{}c@{}}$44.4\pm8.0$\end{tabular} &
  \begin{tabular}[c]{@{}c@{}}$44.6\pm8.2$\end{tabular} &
  \begin{tabular}[c]{@{}c@{}}$48.0\pm13.0$\end{tabular} &
  \begin{tabular}[c]{@{}c@{}}$45.8\pm8.6$\end{tabular} \\ \bottomrule
\end{tabular}
\label{tab:performance}
\end{table*}
\subsection{Deep breath identfication model}
We establish the deep breath identification model based on handcrafted features extracted from processed signals. The kernel function of SVM is set as linear kernel. A total of 15 features are extracted from the 1D temporal signal, including 4 time domain features, 4 short-term domain features, 5 time-frequency domain features, and 2 self-defined features. The first self-defined feature is the standard deviation of the auto-correlation sequence, which represents the stability of respiratory signal. The second self-defined feature is respiratory rate calculated by Welch PSD estimation \cite{welch1967use}. Other features are from \cite{zhao2018noncontact}, which is originally used to detect breathing disorder. 

\section{Experiments and Results}
This section describes the establishment of the breath during walking dataset, and the validation results on this dataset. The performance comparison is shown in \textbf{Tab. \ref{tab:performance}}.
\subsection{Data collection and experimental settings}

A total of 15 healthy subjects (5 female and 10 male) aged from 22 to 39 participated in the experiments. Each subject walked to the depth camera (Azure kinect DK, Microsoft Corporation, Redmond, USA) from a distance of 6 meters, three times with normal breath and three times with deep breath. The speed of walk was depended on themselves. Among 15 subjects, the longest walking time was 18 seconds and the shortest was 6 seconds. During experiments, the wireless piezoelectric respiratory belt (HKH-11L, Huake Electronic Technology Research Institute, Hefei, China) was attached around subjects' abdomen as the ground truth to make sure they performed the correct breath pattern. The depth videos were recorded by PC using kinect recorder tool. Finally, a total of 90 samples were collected, half of which were normal breath and half were deep breath.

\subsection{Results and analysis}

From 15 subjects, every time we randomly choose 10 subjects as training set and the remaining 5 subjects as test set. Then we establish the deep breath identification model based on SVM with the training set using Scikit-learn, and record the performance on test set. We repeat this experiment for 1,000 times, and calculate the mean and standard deviation of the indicators viz. accuracy, precision, recall and F1. The performance of our method is shown in \textbf{Tab. \ref{tab:performance}}. Our method (in bold) yields good performance with the accuracy, precision, recall and F1 of 75.5\%, 76.2\%, 75.0\% and 75.2\%, respectively.

We also do the comparative experiments specific to different signal extraction and processing methods. For signal extraction methods, we compare our method to single ROI method and PCA method. The single ROI method only chooses relative depth between chest and pelvis as breath signal, and the PCA method extracts breath signal from 3D location of breath-related human joints. For signal processing methods, we compare our method to the method which only does band pass filtering. The performance comparison is shown in \textbf{Tab. \ref{tab:performance}}. The results demonstrate that the proposed multiple ROIs extraction in tandem with informative selection achieves the best performance, whether GSA is used or not. Furthermore, the addition of GSA can further obviously improve the model performance. Therefore, the results prove that our signal extraction method and signal processing method have great contributions to the identification of deep breath when someone moves around.

\section{Conclusion}
\label{sec:majhead}
This paper presents a method that can identify deep breath when someone moves forward. Different from previous researches that require the body stay relatively still, we extract the signals from multiple body regions and adopt graph signal analysis for spatial-temporal filtering, so it reduces the noise caused by trunk displacement and deformation. In validation experiments, our method achieves good performance on breath during walking dataset with the accuracy, precision, recall and F1 of 75.5\%, 76.2\%, 75.0\% and 75.2\%, respectively. Besides, the performance comparison demonstrates that our signal extraction method and signal processing method have great contributions to the identification of deep breath while someone moves forward. The proposed approach has significant potential to be applied to the public places to increase the level of service and safety. For example, in a shopping mall, if it is possible to detect the respiratory conditions of customers as they stroll through the mall, and the relevant departments can provide immediate assistance for people who may have or are going through physical or mental trouble.


\clearpage
\bibliography{strings,refs}
\bibliographystyle{IEEEbib}

\end{document}